\documentclass[twocolumn]{IEEEtran}
\usepackage{authblk}
\usepackage{silence}
\usepackage[font=scriptsize]{caption}

\usepackage[pdftex]{graphicx}
\usepackage{amsmath, amsthm, amssymb}
\usepackage[table,xcdraw]{xcolor}
\usepackage{array}
\usepackage{cite}
\usepackage{amssymb}
\usepackage{color}
\usepackage{amsmath}
\usepackage{multirow}
\usepackage{dirtytalk}
\usepackage{amsmath}

\begin{document}

\title{Using Deep Autoencoders for Facial Expression Recognition}
\author[1]{Muhammad Usman}
\author[2,3]{Siddique Latif}
\author[2]{Junaid Qadir}
\affil[1]{COMSATS Institute of Information Technology, Islamabad}
\affil[2]{Information Technology University (ITU), Punjab, Lahore, Pakistan}
\affil[3]{National University of Sciences and Technology (NUST), Islamabad, Pakistan}

\affil[ ]{ {engr.ussman@gmail.com, slatif.msee15seecs@seecs.edu.pk, junaid.qadir@itu.edu.pk}}                
\setcounter{Maxaffil}{0}
\maketitle

\begin{abstract}
Feature descriptors involved in image processing are generally manually chosen and high dimensional in nature. Selecting the most important features is a very crucial task for systems like facial expression recognition. This paper investigates the performance of deep autoencoders for feature selection and dimension reduction for facial expression recognition on multiple levels of hidden layers. The features extracted from the stacked autoencoder outperformed when compared to other state-of-the-art feature selection and dimension reduction techniques.
\end{abstract}

\section{Introduction}

Emotion recognition is an important area of research to enable effective human-computer interaction. Human emotions can be detected using speech signal, facial expressions, body language, and electroencephalography (EEG), etc. In this paper, we focus on facial expression recognition (FER), which is a widely being studied problem \cite{kumari2015facial,fasel2003automatic}. FER has become a very interesting field of study and its applications are not limited to human mental state identification and operator fatigue detection, but also to other scenarios where computers (robots) play a social role such as an instructor, a helper, or even a companion.  In such applications, it is essential that computers are able to recognize human emotions and behave according to their affective states. In healthcare, recognizing patients' emotional instability can help in early diagnosis of psychological disorders \cite{latif20175g}. Another application of FER is to monitor human stress level in daily human-computer interaction.

\begin{figure}[!ht]
\centering
\captionsetup{justification=centering}
\centerline{\includegraphics[width=.5\textwidth]{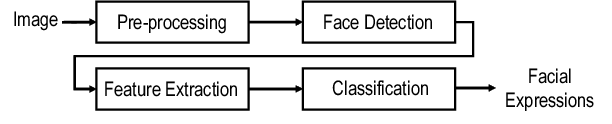}}
\caption{Facial expression recognition (FER) block diagram}
\label{TP}
\end{figure}

Humans can easily recognize another human's emotions using facial expressions but the same task is very challenging for machines. Generally, FER consists of three major steps as shown in the Figure \ref{TP}. The first step involves the detection of a human face from the whole image by using image processing techniques. In the second step, key features are extracted from the detected face. Finally, machine learning models are used to classify images based on the extracted features. 

Features descriptors like histograms of oriented gradients (HOG) \cite{carcagni2015facial}, Local Gabor features \cite{abdulrahman2014gabor} and Weber Local Descriptor (WLD) \cite{wang2013feature} are widely used techniques for FER, whereas HOG has shown to be particularly effective in literature for the task of FER \cite{akinin2015autoencoder}. The dimensionality of these features is usually high. Due to the complexity of multi-view features, dimension reduction and more meaningful representation of this high dimensional data is a challenging task. Therefore, techniques like Principal Component Analysis (PCA) and Local Binary Pattern (LBP),  \cite{liu2017facial,abdulrahman2014gabor}, Non-Negative Matrix Factorization (NMF), etc., are being used to overcome high dimensionality problem by representing the most relevant features in lower-dimensions. 

Machine learning techniques have revolutionized many fields of science including computer vision, pattern recognition, and speech processing through its powerful ability to learn nonlinear relationships over hidden layers, which makes it suitable for automatic features learning and modeling of nonlinear transformations. Deep neural networks (DNNs) can be used for feature extraction as well as for dimensionality reduction \cite{wang2015dimensionality,akinin2015autoencoder}. A large number of classification techniques has been used for FER.  For example, Choi et al. \cite{choi2006realtime} used  artificial neural networks (ANNs) for classification of facial expressions. Authors in \cite{liu2017facial,ghimire2017facial} have used Support Vector Machines (SVMs) for FER. In \cite{siddiqi2013hierarchical,uddin2009enhanced}, authors utilized Hidden Markov Model (HMM) for FER. HMMs are mostly used for frame-level features to handle sequential data. Besides these classifiers, Dynamic Bayesian Networks \cite{li2013simultaneous} and Gaussian Mixture Model \cite{schels2010multiple} are also utilized for learning facial expressions. The recent success of deep learning also motivates its use for FER \cite{liu2014facial,susskind2008generating}.

In this paper, we use a novel approach based on stacked autoencoders for FER. We exploited autoencoders network for effective representation of high dimensional facial features in lower dimensions. Autoencoders represent an ANN configuration in which output units are linked to the input units through the hidden layers. A fewer number of hidden units allow them to represent input data into a low dimensional latent representation. While in stacked autoencoder, output of first layer is immediately given to second layer as an input. In other words, stacked autoencoders are built by stacking additional unsupervised feature learning hidden layers, and can be trained by using greedy methods for each additional layer. As a result, when the data is passed through the multiple hidden layers of stacked autoencoders, it encodes the input vector in a smaller representation more efficiently \cite{liu2015inspired}. In our case, autoencoders network is more suitable as it not only reduces the dimension of data but can also detect most relevant features. In previous work, Hinton et al. \cite{hinton2006reducing} have shown that autoencoders networks can be used for effective dimension reduction and they can produce more effective representation than PCA.

For our experiments, we choose Extended Cohn-Kanade (CK+) \cite{lucey2010extended} dataset which is extensively used for automatic facial image analysis and emotion classification. The HOG features are computed from the selected area of facial expressions and their dimensions have been reduced by using stacked autoencoders on multiple levels and with multiple hidden layers to get the most optimal encoded features. SVM model in the one-vs-all scenario is used for classification on this reduced form of features. We have performed multiple experiments on the selection of optimal dimension (10-500 features) of the feature vector. The feature vector with length 60, obtained after the introduction of four hidden layers in autoencoders network outperformed as compared to other dimensions. Most importantly, we also use PCA for dimension reduction in order to compare the baseline results with autoencoders. Our proposed method for FER using stacked autoencoders is also outperformed when results were compared with PCA and other recent approaches published in this domain. This demonstrates the effectiveness of stacked autoencoders for the selection of the most relevant features for FER task.

The rest of the paper is organized as follows. In Section \ref{Re}, we present background and related work. In Section \ref{pro}, the detail on each step of our proposed method is described. In Section \ref{ER}, we explain the experimental procedure and obtained results. Finally, we conclude in Section \ref{co}.

\section{Related Work}
\label{Re}
Facial expressions are visually observable non-verbal communication signals that occur in response to a person's emotions and originate by the change in facial muscle. They are the key mechanism for conveying and understanding emotions. Ekman and Freisen \cite{ekman1978facial} postulated six universal emotions (i.e., anger, fear, disgust, joy, surprise, and sadness) with distinct content and unique facial expression. Most of the studies in the area of emotion recognition usually focus on classifying these six emotions.

Much of the efforts have been made to classify facial expression with various facial feature by using machine learning algorithms. For example, Anderson et al. \cite{anderson2006real} developed an FER system to recognize the six emotions. They use SVM and Multilayer Perceptrons and achieved a recognition accuracy of 81.82\%. In \cite{wang2013feature}, Wang et al. combined HOG and WLD features to have missing information about the contour and shape. The proposed solution attained 95.86\% recognition rate by using chi-square distance and the nearest neighbor method to classify the fused features. Lia et al. \cite{li2010automatic} used k-nearest neighbor to compare the performance of PCA and NMF on Taiwanese and Indian facial expression databases. They attained above 75\% recognition rate by using both techniques. 

Recently, a comprehensive study has been made by Liu et al. \cite{liu2017facial}, they also combined HOG with Local Binary Patterns (LBP) features. For dimension reduction of extracted features, PCA was used. After applying several classifiers on reduced features, he received 98.3\% maximum recognition rate. Similarly, Xing et al. \cite{xing2016facial}  used local Gabor features with Adaboost classifier for FER and achieved 95.1\% accuracy with the 10-time reduced dimensionality of traditional Gabor features.

Encouragingly, Jung et al. \cite{jung2015joint} used deep neural networks to extract temporal appearance as well as temporal geometric features from RAW data. They tested this technique on several datasets and obtained higher accuracy than state-of-the-art techniques.  Jang et al. \cite{jang2017color} worked on color images and attained 85.74\% recognition rate by using color channel-wise recurrent learning using deep learning. Similarly, Talele et al. \cite{talele2016facial} used LBP features and ANN for classification and the maximum success rate was 95.48\%.

Recently, the autoencoders models have been used more widely for features learning from data and classification problems. For example, Huang et al. \cite{huang2015sparse} used sparse autoencoder networks for feature learning and classification. His technique was good to avoid human interaction but at the cost of computation complexity. Interestingly,  Gupta et al. \cite{gupta2017multi} developed a multi-velocity autoencoder network by using the multi-velocity layers for generating velocity-free deep motion features for facial expressions. The proposed technique attained the state-of-the-art accuracy on various FER datasets. An interesting work has been done by  Makhzan et al. \cite{makhzani2013k} to investigate the effectiveness of sparsity on MNIST data. They showed that sparse autoencoders are simple to train and achieve better classification results as compared to the denoising autoencoders and Restricted Boltzmann Machines (RBMs) as well as networks trained with dropout. Another study \cite{xu2016learning} explored the effect of hidden layers in stacked autoencoders on MNIST data. The authors showed that stacked autoencoders with larger depth have better learning capability but require more training examples and time. 

\section{Proposed Method}
\label{pro}
\begin{figure}[!ht]
\centering
\captionsetup{justification=centering}
\centerline{\includegraphics[width=.4\textwidth]{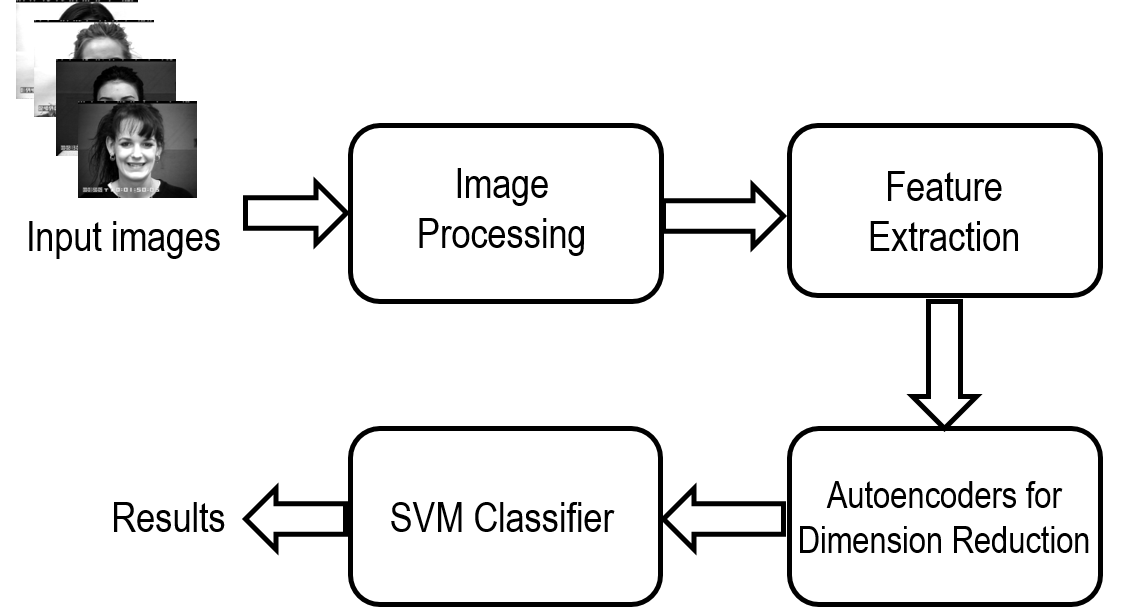}}
\caption{Facial Emotion Recognition (FER) framework}
\label{FC}
\end{figure}

Our proposed FER system consists of four steps (as illustrated in Figure \ref{FC}). The \emph{first step} is related to image processing, in which, we use the state-of-the-art Viola Jones \cite{viola2004robust} face detection method for face detection and extraction. This extracted portion represents the most variance when expression changes. In the \emph{second step}, HOG features are computed from the cropped image. In the \emph{third step}, high-dimensional HOD features are reduced to lower dimension using stacked autoencoders. Finally, in the \emph{fourth step}, the SVM model is used on these lower dimension features to classify the facial expressions. 
Extended Cohn-Kanade Dataset (CK+) is used in our experiment. Most importantly, we investigated the performance of encoded features of length 5 to 100 using different depth of stacked autoencoders. 
Figure. \ref{FC} shows the flowchart of our overall experiment. The detail of each step is given below.

\subsection{Image Processing}
At the image processing stage, we first detect and extract the face region to eliminate redundant regions which can affect recognition rate. The used databases contain much redundant information in the images, and to eliminate the redundant information, the robust real-time detector developed by Viola and Jones \cite{viola2004robust} is employed. In this way, we obtained the face local region around mouth and eyes as these parts represent the most discriminating information when facial expression changes.  
\begin{figure}[!ht]
\centering
\captionsetup{justification=centering}
\centerline{\includegraphics[width=.5\textwidth]{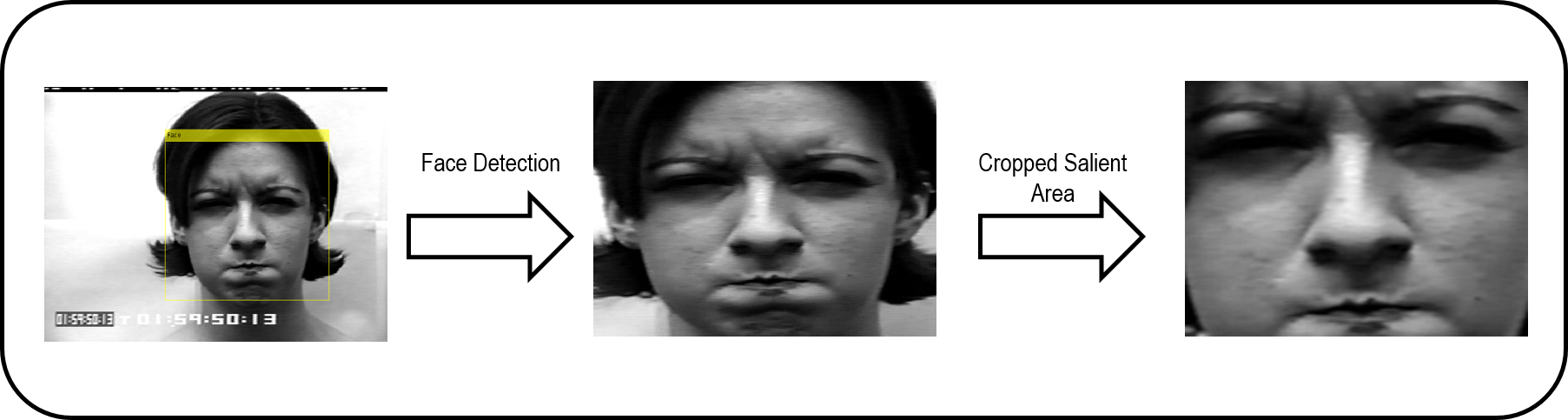}}
\caption{Face detection and cropping salient area}
\label{FD}
\end{figure}

As shown in Figure. \ref{FD}, we crop out the face region and re-size the face image to 128$\times$128 to get the salient areas of facial expression.

\subsection{Feature Extraction}
Histogram of oriented gradients (HOG) is a feature descriptor which is widely used in computer vision and image processing \cite{dalal2005histograms}. The technique counts occurrences of gradient orientation in localized portions of an image. HOG is invariant to geometric and photometric transformations, except for object orientation. The images that are in the database have different expressions and different orientations of eyes, nose, and lip corners. HOG is used in our algorithm because it is a powerful descriptor to detect the variations (i.e., when facial expressions change). In our proposed approach, we applied HOG on cropped face images and extracted the feature vectors. The cropped image of size 128$\times$128 gives a feature vector of size 1$\times$8100 using HOG. The feature vectors are concatenated to form feature matrix as shown in table \ref{Sen}.
\begin{table}[!ht]
\centering
 \begin{tabular}{|c|} 
 \hline
 1x8100 (HOG Feature of image 1) \\ 
 \hline\hline
 1x8100 (HOG Feature of image 2) \\ 
 \hline
 1x8100 (HOG Feature of image 3)\\
 \hline
 1x8100 (HOG Feature of image 4) \\
 \hline
 ............ \\
 \hline
 ............ \\
 \hline
 ............ \\
 \hline
 ............ \\  
 \hline
  
 1x8100 (HOG Feature of image (N-1)) \\
  \hline
 1x8100 (HOG Feature of image N) \\
 \hline
 
\end{tabular}
\caption{HOG feature Matrix of N$\times$8100}
\label{Sen}
\end{table}

\subsection{Dimension Reduction}
The main aim of this work is to show that how nonlinear machine learning technique can be effectively used to obtain the most relevant holistic representation of features in a lower dimension. As the extracted HOG features have a high dimension (N$\times$8100) as compared to the number of available images (327). The state-of-the-art is to reduce the dimension of the features vector by using different dimension reduction techniques such as PCA, linear discriminant analysis (LDA) and NMF. For this purpose, we use autoencoder network on high dimensional feature descriptors extracted by using HOG. To compare the performance of these features, we also use PCA for dimension reduction of features. Both of these techniques are discussed below.

\subsubsection{Autoencoders}

\begin{figure}[!ht]
\centering
\captionsetup{justification=centering}
\centerline{\includegraphics[width=.3\textwidth]{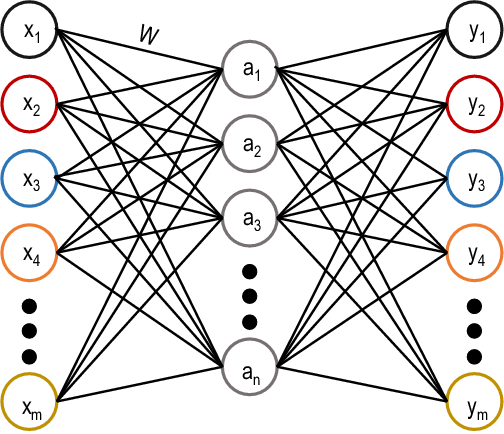}}
\caption{Architecture of an autoencoder network. The input vector $X$ of length $m$  is encoded to lower dimensional feature vector $a$ of length $n$ then reconstructed as $y$ with length $m$ similar to $x$ ($n<m$)}
\label{EN}
\end{figure}

An autoencoder is an unsupervised architecture that replicates the given input at its output. It takes an input feature vector $X$ and learns a code dictionary by changing the raw input data from one representation to another.  An autoencoder applies backpropagation by setting the target values to be equal to the
inputs (i.e., $x^{(i)}=y^{(i)}$) as shown in the Figure \ref{EN}.

For example, if autoencoders are inputted with correlated structural data, then the network will discover some of these correlations \cite{leng20153d}.  In an autoencoder, the lower dimension $a$ is represented by
\begin{equation}
    a=f\left(\sum{Wx+b}\right)
\end{equation}
Where $W$ is associated weight vector with the input unit and hidden unit, $b$ is the bias associated with the hidden unit and $a$ is the activation of the hidden unit in the network. Similarly,  $f(x)$ is the sigmoid function that is given by
\begin{equation}
    f(x)=\frac{1}{1+exp^{(-z)}}
\end{equation}
\begin{equation}
    z_{i}= \sum_{j=1}^{m}W_{ij}x+b_{i}
\end{equation}
and 
\begin{equation}
    a_{i}=f(z_{i})
\end{equation}
The stacked autoencoder can be described as follow
\begin{equation}
     a_{i}=f\left(\sum{W_{i}a_{(i-1)}+b_{i}}\right)
\end{equation}
Encouragingly, an autoencoder can also discover the interesting structure of data, even when the number of hidden units is large, by imposing sparsity constraint on the hidden units. Such architecture is called sparse autoencoder. The cost function $J(W,b)$ of a sparse autoencoder is given by:

\begin{align}
\label{ll}
 J(W,b)= \left[{1\over m}\sum_{i=1}^{m}\left({1\over 2}\vert \vert h_{{\bf W},{\bf b}}({\bf x}^{(i)})-{\bf x}^{(i)}\vert \vert^{2}\right)\right]\notag\\+{\lambda_{1}\over 2}\sum_{l=1}^{L-1}\sum_{i=1}^{s_{l}}\sum_{j=1}^{s_{l+1}}(W_{ji}^{(l)})^{2}+\beta \sum_{j=1}^{s_{2}}KL \left(\rho \vert \vert \hat{\rho}_{j}\right)
\end{align}
Where $h_{W,b}(x)$ is an activation function, $W$ and $b$ are  weights and biases respectively. The first term in equation (\ref{ll}) tries to minimize the difference between the input and output. The second term is the weight decay that avoids over-fitting, where $\lambda$ is the parameter for weight decay, $L$ is the number of layers in autoencoder network, and $s_{l}$ denotes the number of units for the $l^{th}$ layer. Similarly, $W^{(l)}_{ji}$ represents the weight value between the $i^{th}$ unit of layer $l$ and the $j^{th}$ unit of layer $l+1$, and $b^{(l)}_{i}$ is the bias associated with unit $i$ in layer $l+1$. The last term is a sparse penalty term, where $\beta$ controls the weight of this term, and $\rho$ is a sparsity parameter and $KL$ is the Kullback-Leibler ($KL$) divergence that is given by

\begin{align}
    KL \left(\rho \vert \vert \hat{\rho}_{j}\right)=\rho \log {\rho \over \hat{\rho}_{j}}+(1-\rho)\log {1 -\rho \over 1-\hat{\rho}_{j}}
\end{align}

Typically, $\rho$ is set to be a small value close to $0$. $KL$ divergence is a standard function used for measuring the difference between two different distributions.

In this experiment, the extracted features using HOG is inputted to autoencoder network to encode them at the desired level of dimension by limiting the hidden units in hidden layers. The number of hidden layers is always experiential. Therefore, we also tried to explore the effect of an increase in the number of hidden layers for stacked autoencoder. This effect typically used for dimension reduction of input data. We have performed multiple experiments to validate our findings. To get a quality of encoded features from autoencoder, we use backpropagation for fine-tuning of network parameters. Mean square error (MSE) is used as a loss function with 400 epochs.

\subsubsection{Principal Component Analysis}
The research domain of pattern recognition and computer vision is dominated by the extensive use of PCA which is also referred as Karhunen-Loeve expansion \cite{omer2015facial}. PCA is a statistical procedure that uses an orthogonal transformation to convert a set of observations of possibly correlated variables into a set of values of linearly uncorrelated variables called principal components. PCA is an effective method to reduce the feature dimension and has been extensively being applied in FER for dimension reduction of features \cite{liu2017facial,abdulrahman2014gabor}. Therefore,  we chose PCA to compare its performance with nonlinear dimension reduction by autoencoder. High dimension feature matrix (N$\times$8100) is reduced to the multiple numbers of dimension (i.e., 10 to 500) using PCA. 

\subsection{Support Vector Machine}
SVMs are very powerful tool for binary as well as multi-class classification problems. Initially,  SVMs was designed for binary classification that separates the binary class data ($k$=2) with a maximized margin. However, for real-world problems, it is often required to discriminate between data for more than two ($k>$2) categories. Therefore, two representative ensemble schemes exist in SVMs, i.e., one-versus-all and one-versus-one to classify multi-class data. In this experiment, we use SVM in the one-vs-all scenario with a Gaussian kernel function. In one-vs-all scheme, SVM constructs $k$ separate binary classifiers to classify $k$-classes of data. The $m^{th}$ binary classifier is trained by using the data from $m^{th}$ class as the positive example and the remaining $k-1$ number of classes as negative examples. During testing, the class label is predicted by the binary classifier that gives maximum output value. For binary classification task with training data $x_{i}(i=1, 2, 3, . . . . N)$ and corresponding labels $y_{i}=\pm1$, the decision function can be formulated as:
\begin{equation}
    f(x)=sign(w^{T}x+b).
\end{equation}
Where $w^{T}x + b = 0$ denotes a separating hyper-plane,  $w$ is a weight vector normal to the separating hyper-plane and $b$ denotes the bias or offset of the hyper-plane. Following is the region between hyper-planes that is also called margin band.

\begin{equation}\label{5}
    \gamma=\frac{2}{\left|w\right|}
\end{equation}
Finally, choosing the optimal values of $w$ and $b$ is formulated as a optimization problem, where equation \ref{5} is maximized subject to the following constrain:
\begin{equation}
    y_{i}(w^{T}x_{i}+b)\geqslant 1     \forall i
\end{equation}

\section{Experimental Results and Discussion}
\label{ER}
\begin{figure}[!ht]
\centering
\captionsetup{justification=centering}
\centerline{\includegraphics[width=.45\textwidth]{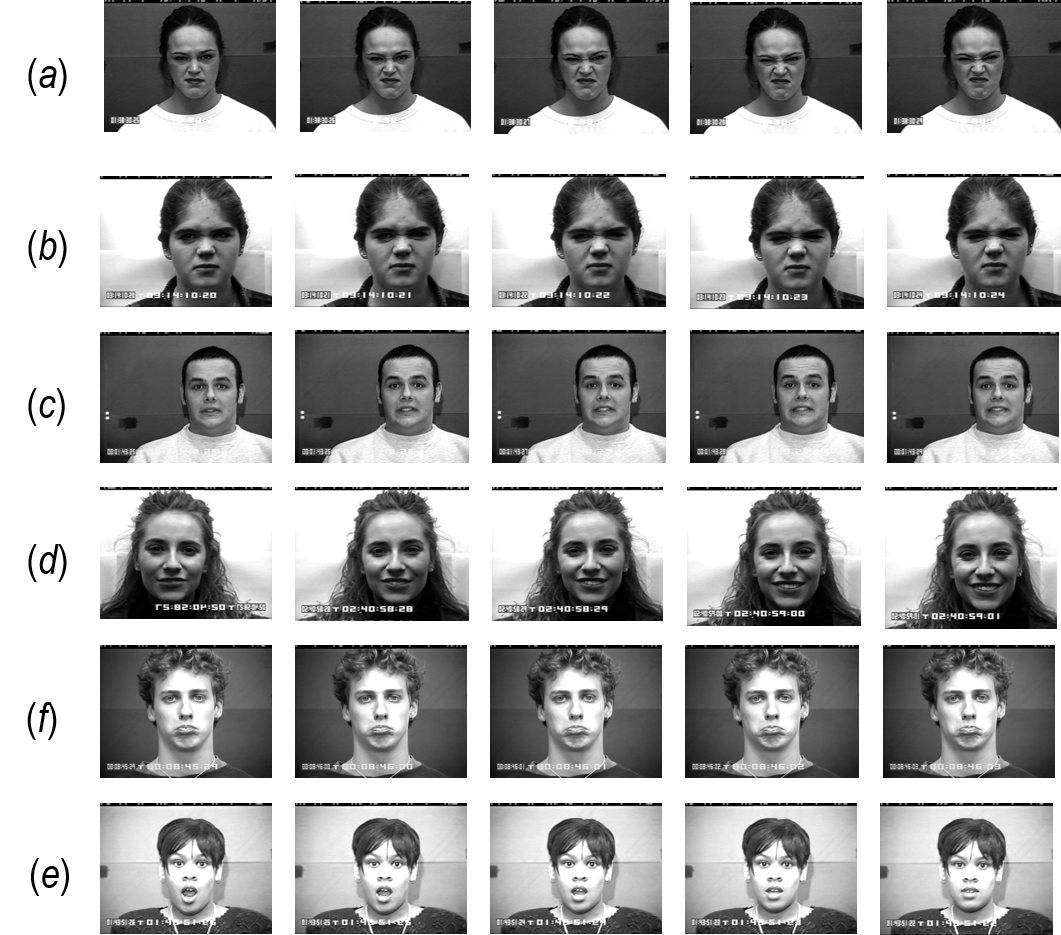}}
\caption{Sample sequence of five images for various emotions: (a) Angry, (b) Disgust, (c) Fear, (d) Happy, (e) Sad and (f) Surprised}
\label{SC}
\end{figure}

The performance of our proposed approach for FER has evaluated on publicly available CK+ database. This dataset contains 593 sequences of images from 123 subjects. Only 327 out of 593 sequences of images are given the labels for 7 human facial expressions. Out of 7 expressions, we used six expressions (i.e., angry, happy, disgust, sadness, surprise, and fear) similar to the methods adopted in \cite{liu2015inspired,zhong2012learning,liu2017facial}. Contempt has only 18 labeled images so it was not included in our experiment. Each expressional image sequence starts with a neutral expression and ends with a peak expression (i.e., anger). In our experiment, we use five peak images of each expression to incorporate the temporal information of an expression. Figure. \ref{SC} shows a sample sequence of images for the six emotions that we use for training. For training, we use 80\% of data while testing was performed using remaining 20\%. During testing, we only use one peak image of each expression.

Multiclass SVM in the one-vs-all scheme with Gaussian kernel has been used for classification of facial expressions using MATLAB. We have performed multiple experiments on a different length of features obtained by autoencoder and PCA as shown in table \ref{ACC}. It can be noted that encoded features obtained by the stacked autoencoders mostly outperformed the baseline (PCA) performance. By using autoencoder for dimension reduction, we achieved the highest recognition rate of 99.60\% with 60 dimensions while with PCA 96.44\% success rate is obtained with 80 dimensions. 

\begin{center}
\begin{table}[!ht]
\centering
\scriptsize

\begin{tabular}{|m{1.2cm}|m{1.7cm} | m{1.6cm}|}
\hline
\textbf{Number of Feature}
& \textbf{PCA (Accuracy \%)}
& \textbf{Autoencoders (Accuracy \%)}
 \\ \hline
\begin{tabular}[c]{@{}l@{}}10\end{tabular}
&\begin{tabular}[c]{@{}l@{}}88.34\end{tabular}
&\begin{tabular}[c]{@{}l@{}}97.80\end{tabular}
\\ \hline
\begin{tabular}[c]{@{}l@{}}20\end{tabular}
&\begin{tabular}[c]{@{}l@{}}90.29\end{tabular}
&\begin{tabular}[c]{@{}l@{}}98.10\end{tabular}
\\ \hline

\begin{tabular}[c]{@{}l@{}}40\end{tabular}
&\begin{tabular}[c]{@{}l@{}}96.01\end{tabular}
&\begin{tabular}[c]{@{}l@{}}98.50\end{tabular}
\\ \hline

\begin{tabular}[c]{@{}l@{}}60\end{tabular}
&\begin{tabular}[c]{@{}l@{}}96.11\end{tabular}
&\begin{tabular}[c]{@{}l@{}}99.60\end{tabular}
\\ \hline
\begin{tabular}[c]{@{}l@{}}80\end{tabular}
&\begin{tabular}[c]{@{}l@{}}96.44\end{tabular}
&\begin{tabular}[c]{@{}l@{}}98.10\end{tabular}
\\ \hline
\begin{tabular}[c]{@{}l@{}}100\end{tabular}
&\begin{tabular}[c]{@{}l@{}}94.17\end{tabular}
&\begin{tabular}[c]{@{}l@{}}98.40\end{tabular}
\\ \hline
\begin{tabular}[c]{@{}l@{}}200\end{tabular}
&\begin{tabular}[c]{@{}l@{}}94.82\end{tabular}
&\begin{tabular}[c]{@{}l@{}}97.84\end{tabular}
\\ \hline
\begin{tabular}[c]{@{}l@{}}300\end{tabular}
&\begin{tabular}[c]{@{}l@{}}95.15\end{tabular}
&\begin{tabular}[c]{@{}l@{}}96.28\end{tabular}
\\ \hline
\begin{tabular}[c]{@{}l@{}}400\end{tabular}
&\begin{tabular}[c]{@{}l@{}}95.46\end{tabular}
&\begin{tabular}[c]{@{}l@{}}96.98\end{tabular}
\\ \hline
\begin{tabular}[c]{@{}l@{}}500\end{tabular}
&\begin{tabular}[c]{@{}l@{}}95.45\end{tabular}
&\begin{tabular}[c]{@{}l@{}}96.98\end{tabular}
\\ \hline

\end{tabular}
\caption{Accuracy using different dimension (number of features) with SVM}
\label{ACC}

\end{table}
\end{center}

\begin{figure}[!ht]
\centering
\captionsetup{justification=centering}
\centerline{\includegraphics[width=.52\textwidth]{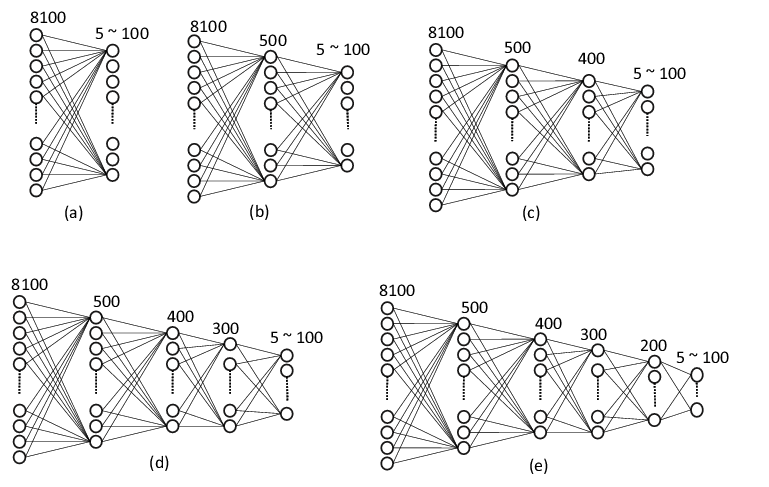}}
\caption{The structure of five stacked autoencoders used in our experiment. (a) one hidden layer autoencoder, (b) two hidden layer autoencoder, (c) three hidden layer autoencoder, (d) four hidden layer autoencoder and (e) five hidden layer autoencoder}
\label{SA}
\end{figure}

We also investigated the effect of adding more hidden layers in autoencoder network. We have performed multiple experiments by introducing more hidden layers with a different number of hidden units (i.e., 500, 400, 300 and 200) while the encoded features are from 5 to 100. Figure \ref{SA} shows the structure of five autoencoders used in our experiments. 

Table \ref{ACC1} shows the results of experiments in which a different number of hidden layers are introduced. It can be noted that higher the number of hidden layers not necessarily increase the accuracy, as already indicated in \cite{chen2014deep,zabalza2016novel}. From these results, we can state that after a certain number of the hidden layer for each number of feature, the accuracy starts decreasing. For example, with 80 features, when hidden layers are introduced, recognition rate increases till the second layer but after that, it decreases.  Similarly, we find the same trend for all number of features but at a different number of hidden layers.

\begin{center}
\begin{table}[!ht]
\centering
\scriptsize

\begin{tabular}{|m{1.2cm}|m{1cm}|m{1cm}|m{1cm}|m{1cm}|m{1cm}|}
\hline
\textbf{Number of Feature}
& \textbf{Hidden Layer 1}
& \textbf{Hidden Layer 2}
& \textbf{Hidden Layer 3}
& \textbf{Hidden Layer 4}
& \textbf{Hidden Layer 5}
\\ \hline
\begin{tabular}[c]{@{}l@{}}5\end{tabular}
&\begin{tabular}[c]{@{}l@{}}78.6\end{tabular}
&\begin{tabular}[c]{@{}l@{}}89.0\end{tabular}
&\begin{tabular}[c]{@{}l@{}}90.1\end{tabular}
&\begin{tabular}[c]{@{}l@{}}93.4\end{tabular}
&\begin{tabular}[c]{@{}l@{}}94.9\end{tabular}
\\ \hline
\begin{tabular}[c]{@{}l@{}}10\end{tabular}
&\begin{tabular}[c]{@{}l@{}}78.4\end{tabular}
&\begin{tabular}[c]{@{}l@{}}93.0\end{tabular}
&\begin{tabular}[c]{@{}l@{}}94.9\end{tabular}
&\begin{tabular}[c]{@{}l@{}}97.8\end{tabular}
&\begin{tabular}[c]{@{}l@{}}97.1\end{tabular}
\\ \hline
\begin{tabular}[c]{@{}l@{}}20\end{tabular}
&\begin{tabular}[c]{@{}l@{}}77.7\end{tabular}
&\begin{tabular}[c]{@{}l@{}}95.2\end{tabular}
&\begin{tabular}[c]{@{}l@{}}98.5\end{tabular}
&\begin{tabular}[c]{@{}l@{}}98.1\end{tabular}
&\begin{tabular}[c]{@{}l@{}}97.8\end{tabular}
\\ \hline
\begin{tabular}[c]{@{}l@{}}40\end{tabular}
&\begin{tabular}[c]{@{}l@{}}96.2\end{tabular}
&\begin{tabular}[c]{@{}l@{}}98.3\end{tabular}
&\begin{tabular}[c]{@{}l@{}}98.9\end{tabular}
&\begin{tabular}[c]{@{}l@{}}98.5\end{tabular}
&\begin{tabular}[c]{@{}l@{}}98.1\end{tabular}
\\ \hline
\begin{tabular}[c]{@{}l@{}}60\end{tabular}
&\begin{tabular}[c]{@{}l@{}}96.9\end{tabular}
&\begin{tabular}[c]{@{}l@{}}98.7\end{tabular}
&\begin{tabular}[c]{@{}l@{}}99.2\end{tabular}
&\begin{tabular}[c]{@{}l@{}}99.6\end{tabular}
&\begin{tabular}[c]{@{}l@{}}98.8\end{tabular}
\\ \hline
\begin{tabular}[c]{@{}l@{}}80\end{tabular}
&\begin{tabular}[c]{@{}l@{}}97.6\end{tabular}
&\begin{tabular}[c]{@{}l@{}}98.9\end{tabular}
&\begin{tabular}[c]{@{}l@{}}98.4\end{tabular}
&\begin{tabular}[c]{@{}l@{}}98.1\end{tabular}
&\begin{tabular}[c]{@{}l@{}}97.3\end{tabular}
\\ \hline
\begin{tabular}[c]{@{}l@{}}100\end{tabular}
&\begin{tabular}[c]{@{}l@{}}97.9\end{tabular}
&\begin{tabular}[c]{@{}l@{}}98.5\end{tabular}
&\begin{tabular}[c]{@{}l@{}}98.9\end{tabular}
&\begin{tabular}[c]{@{}l@{}}98.4\end{tabular}
&\begin{tabular}[c]{@{}l@{}}98.2\end{tabular}
\\ \hline
\end{tabular}
\caption{Recognition rate achieved on multiple hidden layers for different dimension (number of features)}
\label{ACC1}

\end{table}
\end{center}
 
\begin{figure}[!ht]
\centering
\captionsetup{justification=centering}
\centerline{\includegraphics[width=0.45\textwidth]{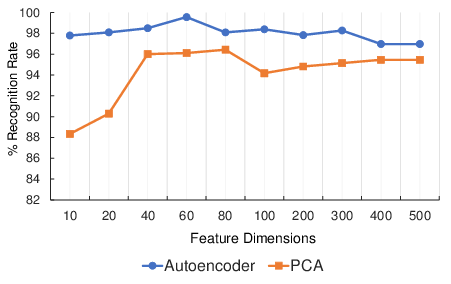}}
\caption{Obtained accuracy by using different dimension of features}
\label{RS}
\end{figure}

Figure. \ref{RS} shows the trend of best recognition rate and the number of reduced dimensions with autoencoder and PCA. It can be seen from Figure. \ref{RS} that there is no regular relationship between accuracy and the number of dimensions, however, it remains almost same after the specific dimension, i.e., 200. 
The accuracy of 99.60 with less (i.e., 60) number of features is not reported in the literature. We have also revived the latest papers in table \ref{CO}, to compare our method with recently published papers in this domain. It can be seen from table \ref{CO}, previously maximum achieved accuracy is 99.51\% using a combination of features (i.e, HOG+LDA+PCA). Similarly, Liu et al. \cite{liu2017facial} achieved 98.3\% recognition rate using local binary patterns (LBP) and HOG features together. They achieved this accuracy using a combination of features with 80 dimensions. While our proposed method has shown significantly better results while using a single type of features at lower dimensions.

\begin{table}[!ht]
\centering
\scriptsize
\begin{tabular}{ |m{2cm} | m{1cm}|m{2.5cm} |m{1.5cm}|}
\hline
\textbf{Study}
& \textbf{Year}
& \textbf{Method}
& \textbf{Accuracy (\%)} 
 \\ \hline
\begin{tabular}[c]{@{}l@{}}Xing et al. \cite{xing2016facial}\end{tabular}
&\begin{tabular}[c]{@{}l@{}}2016\end{tabular}
&\begin{tabular}[c]{@{}l@{}}Local Gabor Feature + \\Adaboost Classifier\end{tabular}
&\begin{tabular}[c]{@{}l@{}}95.1\end{tabular}
\\ \hline
\begin{tabular}[c]{@{}l@{}}Cossetin et al. \cite{cossetin2016facial}\end{tabular}
&\begin{tabular}[c]{@{}l@{}}2016\end{tabular}
&\begin{tabular}[c]{@{}l@{}}Pairwise Feature\end{tabular}
&\begin{tabular}[c]{@{}l@{}}98.07\end{tabular}
\\ \hline
\begin{tabular}[c]{@{}l@{}}Kar et al. \cite{kar2017face}\end{tabular}
&\begin{tabular}[c]{@{}l@{}}2017\end{tabular}
&\begin{tabular}[c]{@{}l@{}}HOG+LDA+PCA
\end{tabular}
&\begin{tabular}[c]{@{}l@{}}99.51\end{tabular}
\\\hline
\begin{tabular}[c]{@{}l@{}}Liu et al. \cite{xing2016facial}\end{tabular}
&\begin{tabular}[c]{@{}l@{}}2017\end{tabular}
&\begin{tabular}[c]{@{}l@{}}Local binary patterns \\(LBP)+HOG with PCA\end{tabular}
&\begin{tabular}[c]{@{}l@{}}98.3\end{tabular}
\\\hline
\begin{tabular}[c]{@{}l@{}}Our study\end{tabular}
&\begin{tabular}[c]{@{}l@{}}2017\end{tabular}
&\begin{tabular}[c]{@{}l@{}}HOG + Autoencoders 
\end{tabular}
&\begin{tabular}[c]{@{}l@{}}99.60\end{tabular}
\\\hline
\end{tabular}
\caption{Comparison of some recent paper with our study}
\label{CO}
\end{table}

Although our proposed approach has achieved a state-of-the-art recognition rate but the time complexity of autoencoders is linearly dependent on the number of features and hidden layers. Greater the number of features or hidden layers, the more time that is required to train the model.

\section{Conclusion}
\label{co}
The main contribution of this paper is to investigate the performance of deep autoencoders for lower dimensional feature representation. The experiment proves that nonlinear dimension reduction using autoencoders is more effective than linear dimension reduction techniques for FER. We used CK+ dataset in our experiments and compared our results using features obtained by autoencoder networks with state-of-the-art PCA. Most importantly, we explored the effect of an increase in the number of hidden layers which enhanced the learning capability of the network to provide more robust and optimal features for facial expression recognition.


\begin{thebibliography}{10}
\providecommand{\url}[1]{#1}
\csname url@samestyle\endcsname
\providecommand{\newblock}{\relax}
\providecommand{\bibinfo}[2]{#2}
\providecommand{\BIBentrySTDinterwordspacing}{\spaceskip=0pt\relax}
\providecommand{\BIBentryALTinterwordstretchfactor}{4}
\providecommand{\BIBentryALTinterwordspacing}{\spaceskip=\fontdimen2\font plus
\BIBentryALTinterwordstretchfactor\fontdimen3\font minus
  \fontdimen4\font\relax}
\providecommand{\BIBforeignlanguage}[2]{{%
\expandafter\ifx\csname l@#1\endcsname\relax
\typeout{** WARNING: IEEEtran.bst: No hyphenation pattern has been}%
\typeout{** loaded for the language `#1'. Using the pattern for}%
\typeout{** the default language instead.}%
\else
\language=\csname l@#1\endcsname
\fi
#2}}
\providecommand{\BIBdecl}{\relax}
\BIBdecl

\bibitem{kumari2015facial}
J.~Kumari, R.~Rajesh, and K.~Pooja, ``Facial expression recognition: A
  survey,'' \emph{Procedia Computer Science}, vol.~58, pp. 486--491, 2015.

\bibitem{fasel2003automatic}
B.~Fasel and J.~Luettin, ``Automatic facial expression analysis: a survey,''
  \emph{Pattern recognition}, vol.~36, no.~1, pp. 259--275, 2003.

\bibitem{latif20175g}
S.~Latif, J.~Qadir, S.~Farooq, and M.~A. Imran, ``How 5g (and concomitant
  technologies) will revolutionize healthcare,'' \emph{arXiv preprint
  arXiv:1708.08746}, 2017.

\bibitem{carcagni2015facial}
P.~Carcagn{\`\i}, M.~Del~Coco, M.~Leo, and C.~Distante, ``Facial expression
  recognition and histograms of oriented gradients: a comprehensive study,''
  \emph{SpringerPlus}, vol.~4, no.~1, p. 645, 2015.

\bibitem{abdulrahman2014gabor}
M.~Abdulrahman, T.~R. Gwadabe, F.~J. Abdu, and A.~Eleyan, ``Gabor wavelet
  transform based facial expression recognition using pca and lbp,'' in
  \emph{Signal Processing and Communications Applications Conference (SIU),
  2014 22nd}.\hskip 1em plus 0.5em minus 0.4em\relax IEEE, 2014, pp.
  2265--2268.

\bibitem{wang2013feature}
X.~Wang, C.~Jin, W.~Liu, M.~Hu, L.~Xu, and F.~Ren, ``Feature fusion of hog and
  wld for facial expression recognition,'' in \emph{System Integration (SII),
  2013 IEEE/SICE International Symposium on}.\hskip 1em plus 0.5em minus
  0.4em\relax IEEE, 2013, pp. 227--232.

\bibitem{akinin2015autoencoder}
M.~V. Akinin, N.~V. Akinina, A.~I. Taganov, and M.~B. Nikiforov, ``Autoencoder:
  Approach to the reduction of the dimension of the vector space with
  controlled loss of information,'' in \emph{Embedded Computing (MECO), 2015
  4th Mediterranean Conference on}.\hskip 1em plus 0.5em minus 0.4em\relax
  IEEE, 2015, pp. 171--173.

\bibitem{liu2017facial}
Y.~Liu, Y.~Li, X.~Ma, and R.~Song, ``Facial expression recognition with fusion
  features extracted from salient facial areas,'' \emph{Sensors}, vol.~17,
  no.~4, p. 712, 2017.

\bibitem{wang2015dimensionality}
Y.~Wang, H.~Yao, S.~Zhao, and Y.~Zheng, ``Dimensionality reduction strategy
  based on auto-encoder,'' in \emph{Proceedings of the 7th International
  Conference on Internet Multimedia Computing and Service}.\hskip 1em plus
  0.5em minus 0.4em\relax ACM, 2015, p.~63.

\bibitem{choi2006realtime}
H.-C. Choi and S.-Y. Oh, ``Realtime facial expression recognition using active
  appearance model and multilayer perceptron,'' in \emph{SICE-ICASE, 2006.
  International Joint Conference}.\hskip 1em plus 0.5em minus 0.4em\relax IEEE,
  2006, pp. 5924--5927.

\bibitem{ghimire2017facial}
D.~Ghimire, S.~Jeong, J.~Lee, and S.~H. Park, ``Facial expression recognition
  based on local region specific features and support vector machines,''
  \emph{Multimedia Tools and Applications}, vol.~76, no.~6, pp. 7803--7821,
  2017.

\bibitem{siddiqi2013hierarchical}
M.~H. Siddiqi, S.~Lee, Y.-K. Lee, A.~M. Khan, and P.~T.~H. Truc, ``Hierarchical
  recognition scheme for human facial expression recognition systems,''
  \emph{Sensors}, vol.~13, no.~12, pp. 16\,682--16\,713, 2013.

\bibitem{uddin2009enhanced}
M.~Z. Uddin, J.~Lee, and T.-S. Kim, ``An enhanced independent component-based
  human facial expression recognition from video,'' \emph{IEEE Transactions on
  Consumer Electronics}, vol.~55, no.~4, 2009.

\bibitem{li2013simultaneous}
Y.~Li, S.~Wang, Y.~Zhao, and Q.~Ji, ``Simultaneous facial feature tracking and
  facial expression recognition,'' \emph{IEEE Transactions on Image
  Processing}, vol.~22, no.~7, pp. 2559--2573, 2013.

\bibitem{schels2010multiple}
M.~Schels and F.~Schwenker, ``A multiple classifier system approach for facial
  expressions in image sequences utilizing gmm supervectors,'' in \emph{Pattern
  Recognition (ICPR), 2010 20th International Conference on}.\hskip 1em plus
  0.5em minus 0.4em\relax IEEE, 2010, pp. 4251--4254.

\bibitem{liu2014facial}
P.~Liu, S.~Han, Z.~Meng, and Y.~Tong, ``Facial expression recognition via a
  boosted deep belief network,'' in \emph{Proceedings of the IEEE Conference on
  Computer Vision and Pattern Recognition}, 2014, pp. 1805--1812.

\bibitem{susskind2008generating}
J.~M. Susskind, G.~E. Hinton, J.~R. Movellan, and A.~K. Anderson, ``Generating
  facial expressions with deep belief nets,'' in \emph{Affective
  Computing}.\hskip 1em plus 0.5em minus 0.4em\relax InTech, 2008.

\bibitem{liu2015inspired}
M.~Liu, S.~Li, S.~Shan, and X.~Chen, ``Au-inspired deep networks for facial
  expression feature learning,'' \emph{Neurocomputing}, vol. 159, pp. 126--136,
  2015.

\bibitem{hinton2006reducing}
G.~E. Hinton and R.~R. Salakhutdinov, ``Reducing the dimensionality of data
  with neural networks,'' \emph{science}, vol. 313, no. 5786, pp. 504--507,
  2006.

\bibitem{lucey2010extended}
P.~Lucey, J.~F. Cohn, T.~Kanade, J.~Saragih, Z.~Ambadar, and I.~Matthews, ``The
  extended cohn-kanade dataset (ck+): A complete dataset for action unit and
  emotion-specified expression,'' in \emph{Computer Vision and Pattern
  Recognition Workshops (CVPRW), 2010 IEEE Computer Society Conference
  on}.\hskip 1em plus 0.5em minus 0.4em\relax IEEE, 2010, pp. 94--101.

\bibitem{ekman1978facial}
P.~Ekman and W.~Friesen, ``Facial action coding system: a technique for the
  measurement of facial movement,'' \emph{Palo Alto: Consulting Psychologists},
  1978.

\bibitem{anderson2006real}
K.~Anderson and P.~W. McOwan, ``A real-time automated system for the
  recognition of human facial expressions,'' \emph{IEEE Transactions on
  Systems, Man, and Cybernetics, Part B (Cybernetics)}, vol.~36, no.~1, pp.
  96--105, 2006.

\bibitem{li2010automatic}
J.~Li and M.~Oussalah, ``Automatic face emotion recognition system,'' in
  \emph{Cybernetic Intelligent Systems (CIS), 2010 IEEE 9th International
  Conference on}.\hskip 1em plus 0.5em minus 0.4em\relax IEEE, 2010, pp. 1--6.

\bibitem{xing2016facial}
Y.~Xing and W.~Luo, ``Facial expression recognition using local gabor features
  and adaboost classifiers,'' in \emph{Progress in Informatics and Computing
  (PIC), 2016 International Conference on}.\hskip 1em plus 0.5em minus
  0.4em\relax IEEE, 2016, pp. 228--232.

\bibitem{jung2015joint}
H.~Jung, S.~Lee, J.~Yim, S.~Park, and J.~Kim, ``Joint fine-tuning in deep
  neural networks for facial expression recognition,'' in \emph{Proceedings of
  the IEEE International Conference on Computer Vision}, 2015, pp. 2983--2991.

\bibitem{jang2017color}
J.~Jang, D.~H. Kim, H.-I. Kim, and Y.~M. Ro, ``Color channel-wise recurrent
  learning for facial expression recognition,'' in \emph{Acoustics, Speech and
  Signal Processing (ICASSP), 2017 IEEE International Conference on}.\hskip 1em
  plus 0.5em minus 0.4em\relax IEEE, 2017, pp. 1233--1237.

\bibitem{talele2016facial}
K.~Talele, A.~Shirsat, T.~Uplenchwar, and K.~Tuckley, ``Facial expression
  recognition using general regression neural network,'' in \emph{Bombay
  Section Symposium (IBSS), 2016 IEEE}.\hskip 1em plus 0.5em minus 0.4em\relax
  IEEE, 2016, pp. 1--6.

\bibitem{huang2015sparse}
B.~Huang and Z.~Ying, ``Sparse autoencoder for facial expression recognition,''
  in \emph{Ubiquitous Intelligence and Computing and 2015 IEEE 12th Intl Conf
  on Autonomic and Trusted Computing and 2015 IEEE 15th Intl Conf on Scalable
  Computing and Communications and Its Associated Workshops (UIC-ATC-ScalCom),
  2015 IEEE 12th Intl Conf on}.\hskip 1em plus 0.5em minus 0.4em\relax IEEE,
  2015, pp. 1529--1532.

\bibitem{gupta2017multi}
O.~Gupta, D.~Raviv, and R.~Raskar, ``Multi-velocity neural networks for facial
  expression recognition in videos,'' \emph{IEEE Transactions on Affective
  Computing}, 2017.

\bibitem{makhzani2013k}
A.~Makhzani and B.~Frey, ``K-sparse autoencoders,'' \emph{arXiv preprint
  arXiv:1312.5663}, 2013.

\bibitem{xu2016learning}
Q.~Xu, C.~Zhang, L.~Zhang, and Y.~Song, ``The learning effect of different
  hidden layers stacked autoencoder,'' in \emph{Intelligent Human-Machine
  Systems and Cybernetics (IHMSC), 2016 8th International Conference on},
  vol.~2.\hskip 1em plus 0.5em minus 0.4em\relax IEEE, 2016, pp. 148--151.

\bibitem{viola2004robust}
P.~Viola and M.~J. Jones, ``Robust real-time face detection,''
  \emph{International journal of computer vision}, vol.~57, no.~2, pp.
  137--154, 2004.

\bibitem{dalal2005histograms}
N.~Dalal and B.~Triggs, ``Histograms of oriented gradients for human
  detection,'' in \emph{Computer Vision and Pattern Recognition, 2005. CVPR
  2005. IEEE Computer Society Conference on}, vol.~1.\hskip 1em plus 0.5em
  minus 0.4em\relax IEEE, 2005, pp. 886--893.

\bibitem{leng20153d}
B.~Leng, S.~Guo, X.~Zhang, and Z.~Xiong, ``3d object retrieval with stacked
  local convolutional autoencoder,'' \emph{Signal Processing}, vol. 112, pp.
  119--128, 2015.

\bibitem{omer2015facial}
A.~E. Omer and A.~Khurran, ``Facial recognition using principal component
  analysis based dimensionality reduction,'' in \emph{Computing, Control,
  Networking, Electronics and Embedded Systems Engineering (ICCNEEE), 2015
  International Conference on}.\hskip 1em plus 0.5em minus 0.4em\relax IEEE,
  2015, pp. 434--439.

\bibitem{zhong2012learning}
L.~Zhong, Q.~Liu, P.~Yang, B.~Liu, J.~Huang, and D.~N. Metaxas, ``Learning
  active facial patches for expression analysis,'' in \emph{Computer Vision and
  Pattern Recognition (CVPR), 2012 IEEE Conference on}.\hskip 1em plus 0.5em
  minus 0.4em\relax IEEE, 2012, pp. 2562--2569.

\bibitem{chen2014deep}
Y.~Chen, Z.~Lin, X.~Zhao, G.~Wang, and Y.~Gu, ``Deep learning-based
  classification of hyperspectral data,'' \emph{IEEE Journal of Selected topics
  in applied earth observations and remote sensing}, vol.~7, no.~6, pp.
  2094--2107, 2014.

\bibitem{zabalza2016novel}
J.~Zabalza, J.~Ren, J.~Zheng, H.~Zhao, C.~Qing, Z.~Yang, P.~Du, and
  S.~Marshall, ``Novel segmented stacked autoencoder for effective
  dimensionality reduction and feature extraction in hyperspectral imaging,''
  \emph{Neurocomputing}, vol. 185, pp. 1--10, 2016.

\bibitem{cossetin2016facial}
M.~J. Cossetin, J.~C. Nievola, and A.~L. Koerich, ``Facial expression
  recognition using a pairwise feature selection and classification approach,''
  in \emph{Neural Networks (IJCNN), 2016 International Joint Conference
  on}.\hskip 1em plus 0.5em minus 0.4em\relax IEEE, 2016, pp. 5149--5155.

\bibitem{kar2017face}
N.~B. Kar, K.~S. Babu, and S.~K. Jena, ``Face expression recognition using
  histograms of oriented gradients with reduced features,'' in
  \emph{Proceedings of International Conference on Computer Vision and Image
  Processing}.\hskip 1em plus 0.5em minus 0.4em\relax Springer, 2017, pp.
  209--219.

\end{thebibliography}


\end{document}